\documentclass[final,5p,times,twocolumn]{elsarticle}

\usepackage{lineno}
\usepackage{graphicx}
\usepackage{amssymb}
\usepackage{pdflscape}
\usepackage[english]{babel}
\usepackage[dvipsnames,svgnames,x11names]{xcolor}
\usepackage{booktabs,tabularx}
\usepackage{multirow}
\usepackage{subfig} 
\usepackage{hyperref}
\usepackage{amsmath}
\usepackage{commath}
\usepackage[english]{babel}
\usepackage[ruled]{algorithm2e}
\SetEndCharOfAlgoLine{}
\usepackage[leftmargin=6em,rightmargin=12em,indentfirst=false]{quoting}
\usepackage{siunitx}

\usepackage[final]{changes}
\usepackage{float}

\usepackage{lineno}
\usepackage{tikz}
\usepackage[export]{adjustbox}

\usepackage{graphicx}
\usepackage[utf8]{inputenc}
\usepackage[export]{adjustbox}
\usepackage{wrapfig}
\usepackage{dcolumn}

\makeatletter
\newcolumntype{T}[3]{>{\textfont0=\the@{#1}{#2}{#3}}c<{\DC@end}}
\makeatother

\usepackage{pgfplots}
\pgfplotsset{width=10cm,compat=1.9}

\usepackage{array}

\newcolumntype{L}[1]{>{\raggedright\let\newline\\\arraybackslash\hspace{0pt}}m{#1}}
\newcolumntype{C}[1]{>{\centering\let\newline\\\arraybackslash\hspace{0pt}}m{#1}}
\newcolumntype{R}[1]{>{\raggedleft\let\newline\\\arraybackslash\hspace{0pt}}m{#1}}

\usepackage{subfiles}

\usepackage{todonotes}

\journal{Building and Environment}
\begin{document}
	
\begin{frontmatter}

\title{Targeting occupant feedback using digital twins: Adaptive spatial-temporal thermal preference sampling to optimize personal comfort models}

\author{Mahmoud M. Abdelrahman, Clayton Miller$^{*}$}

\address{Department of the Built Environment, College of Design and Engineering, National University of Singapore (NUS), Singapore}
\address{$^*$Corresponding Author: clayton@nus.edus.sg, +65 81602452}

\begin{abstract}
Collecting intensive longitudinal thermal preference data from building occupants is emerging as an innovative means of characterizing the performance of buildings and the people who use them. These techniques have occupants giving subjective feedback using smartphones or smartwatches frequently over the course of days or weeks. The intention is that the data will be collected with high spatial and temporal diversity to best characterize a building and the occupant's preferences. But in reality, leaving the occupant to respond in an ad-hoc or fixed interval way creates unneeded survey fatigue and redundant data. This paper outlines a scenario-based (virtual experiment) method for optimizing data sampling using a smartwatch to achieve comparable accuracy in a personal thermal preference model with fewer data. This method uses BIM-extracted spatial data and Graph Neural Network-based (GNN) modeling to find regions of similar comfort preference to identify the best scenarios for triggering the occupant to give feedback. This method is compared to two baseline scenarios that use conventional zoning and a generic 4x4 square meter grid method from two field-based data sets. The results show that the proposed Build2Vec method has an 18-23\% higher overall sampling quality than the spaces-based and square-grid-based sampling methods. The Build2Vec method also performs similar to the baselines when removing redundant occupant feedback points but with better scalability potential.
\end{abstract}


\end{frontmatter}


\section{Introduction}
\label{sec:introduction}

The focus on predicting personal thermal comfort in the built environment has grown over the last decade to improve occupant satisfaction. Several recent vital studies have shown promise in capturing individual differences in comfort preference, but there are still deficiencies to be addressed. The literature indicates that conventional thermal comfort modeling approaches~\cite{In,VanHoof2008} that use the Predicted Mean Vote (PMV) have been shown to be correct only 34 percent of the time~\cite{Cheung2019}. Another study shows a significant gap in delivering at least one or two factors that impact indoor environmental satisfaction~\cite{Asadi2017}. Mainly, this gap occurs because people are physio-psychologically different from each other, buildings are spatially diverse, and one set of comfort conditions does not fit all occupants~\cite{Kim2013, Lahtinen2002PsychosocialProblem, Guenther2019-hb}. 

Therefore, longitudinally intensive field-based data collection methods are emerging in which the goal is to characterize the \emph{personal} tendencies to feel thermal comfort sensation and preference~\cite{Arakawa_Martins2022-vh, Aryal2020-tp}. These individualized models began with work collecting conventional environmental sensor data and longitudinal data from occupants for models to outperform PMV~\cite{Kim2018-yh, Zhao2014-ja, Pazhoohesh2018-ax} and grew to include data collection from wearable~\cite{Liu2019-pi, Katic2020-pt} and infrared sensors~\cite{Li2018-hz, Aryal2019-bx}. The applications for these models grew to include energy management~\cite{Konis2017-eo, ABDELRAHMAN2021110885}, personalized profiles~\cite{Lee2017-tx}, and HVAC control~\cite{Lee2020-wn,Li2017-fh, Jung2019-nt, Aguilera2019-ma}. Through most of these studies, personal thermal comfort data from individuals were collected using computer and smartphone interfaces. To decrease survey fatigue and increase spatial and temporal diversity, smartwatches were used to collect data using micro-ecological momentary assessments (EMA) from building occupants about their satisfaction perception momentarily within certain environmental conditions~\cite{Jayathissa2020-pv, Jayathissa2019-kg, Sae-Zhang2020-fy}. 


Despite the advancements in the ability to collect more data longitudinally from occupants, there is still the challenge of optimizing the \emph{where} and \emph{when} that sampling should occur. Collecting many comfort preference data points from an occupant is of little use if the feedback is taken from the same space at the same conditions. The comfort preference prediction models need a diversity of feedback samples from a range of spaces and conditions in the built environment context to achieve the best generalizable performance.

Another challenge is the selection and placement of indoor environmental sensors. The collection of these data has many challenges related to sensor spatial-temporal resolution, cost and accuracy, sampling frequency, and method selection. The location distribution of sensors in a space is referred to as \textit{spatial resolution}. In contrast, the time interval at which data is measured is referred to as \textit{temporal resolution} (e.g., every one second). Some previous literature includes guidelines on the spatial and temporal resolution~\cite{Heinzerling2013}.

\begin{table*}
	\centering
	\caption{Examples of IEQ Spatial-temporal resolution guidelines from literature.}
	\begin{tabular}{|p{0.09\linewidth}|p{0.03\linewidth}|p{0.1\linewidth}|p{0.15\linewidth}|p{0.12\linewidth}|p{0.12\linewidth}|p{0.15\linewidth}|}
		\hline
		\textbf{\tiny{Protocol}}                                  & \textbf{\tiny{Ref}}   & \textbf{\tiny{Resolution}} & \textbf{\tiny{Thermal}}                                                                                              & \textbf{\tiny{Acoustics}}                                                 & \textbf{\tiny{Lighting}}                                                                     & \textbf{\tiny{IAQ}}                                                                                      \\
		\hline
		\tiny{Performance Measurements Protocol (PMP)}            &   \tiny{\cite{Haberl2008-sz}}       & \textbf{\tiny{Spatial}}    & \tiny{Close to locations where occupants identified issues; Areas of control system (diffusers, radiators, windows)} & \tiny{At least 4 locations per occupied room}                             & \tiny{Regular grid spacing = 0.25 space between luminaires Luminance: 11 specific locations} & \tiny{Spaces with unusual or atypical activities; omit sparsely occupied and unoccupied areas}           \\
		                                                          &                       & \textbf{\tiny{Temporal}}   & \tiny{Continuous for unknown length of time}                                                                         & \tiny{Background noise: 30 seconds minimum per measurement}               & {\_\_}                                                                                       & \tiny{Continuous for at least 1 week}                                                                    \\

		\hline
		\tiny{The US. Environmental Protection Agency (EPA BASE)} & \tiny{\cite{Epa2003}} & \textbf{\tiny{Spatial}}    & \tiny{At the height of 1.1 m above the floor at three fixed indoor sites}                                            & \tiny{At the height of 1.1 m above the floor at three fixed indoor sites} & \tiny{At the height of 1.1 m above the floor at three fixed indoor sites}                    & \tiny{Divide the study area into 5m x 5m tiles ;Randomly select tiles that define monitoring locations } \\
		                                                          &                       & \textbf{\tiny{Temporal}}   & \tiny{Continuous over three-day period}                                                                              & \tiny{Continuous over three-day period}                                   & \tiny{Continuous over three-day period}                                                      & \tiny{Continuous over three-day period}                                                                  \\

		\hline
	\end{tabular}%
	\label{tab:spatial_temporal_resolution}%
\end{table*}

Table \ref{tab:spatial_temporal_resolution} outlines two guidelines from the literature in the area of spatial-temporal sampling. It is observed that such guidelines are fuzzy as they do not give exhaustive guidance for all combinations of space settings. The implementation of these guidelines does not always capture the heterogeneous nature of spaces. Furthermore, the temporal resolution is primarily taken in a fixed time interval, e.g., \emph{30 minutes}. Built environment spaces are known to be significantly heterogeneous \cite{TARABIEH2019311}. For example, the perception of thermal comfort can differ significantly within a few meters due to situations like exposure to air movement from a ceiling fan or solar radiation next to the window. The influence of spatial objects their nearby elements, known as the Area-of-Influence (AoI), has been studied mainly in literature and is referred to as \emph{spatial proximity} or \emph{spatial nearness}.

\subsection{Targeted occupant survey techniques to improve data collection}
In order to address the deficiencies in strategically collecting sensor and feedback information from the built environment, a previous study introduced the concept of \emph{Targeted Occupant Survey (TOS)} as an adaptive \emph{right-now} survey where questions are triggered only upon need, e.g., when a specific environmental condition is met~\cite{DuarteRoa2020TargetedConditions}. This approach was shown to be efficient in lessening survey fatigue and decreasing the data storage needs. More importantly, the method helps produce a balanced dataset; that is, the survey stops when it reaches a pre-defined minimum number of data inputs from a particular location or combination of conditions. However, there are some limitations in using the TOS technique as it requires installing sensors at every workstation. These sensors have a cost of deployment, and they need a certain level of accuracy, calibration, and maintenance. The approach relies on the sensors to be WiFi-connected and send data in real-time to the database to trigger the survey. 

\subsection{Using digital twins to optimize spatial and temporal sampling of subjective feedback}

The approach outlined in this paper builds upon the TOS technique and work in graph-based modeling for thermal comfort~\cite{Abdelrahman2022-aj} by including the ability to leverage digital twin data from Building Information Models (BIM). This method focuses on the increase of flexibility and scalability of determining \emph{where} and \emph{when} data is collected from occupants. The definition of a digital twin from the literature is that it's a virtual representation of an object or system that spans its lifecycle, is updated from real-time data, and uses simulation, machine learning, and reasoning to help decision-making~\cite{Grieves2017-li}. This research introduces a novel method for sampling occupants' satisfaction feedback when a spatial-temporal triggering condition is met. This approach aligns with the digital twin definition as there is a virtual model of the spaces extracted from a BIM that informs the optimal sampling rate of a real-world data collection process~\cite{Miller2021-bm}. This method can be done continuously and in real-time if needed and aims to leverage buildings' spatial data and user location to scale data collection without adding more sensors. It is built on a model called Build2Vec~\cite{abdelrahmanbuild2vec} that is used to extract similar spatial locations in buildings and differentiate distinct (heterogeneous) zones. The similarity concept allows using the same constant number of zones when expanding the experiment (adding new spaces or new users). 


\subsection{Research objectives}
This research aims to optimize the thermal comfort subjective data sampling in a building using spatial proximity to different building elements by: 
\begin{enumerate}
    \item Minimizing the number of feedback votes (to reduce survey fatigue); 
    \item Minimizing the number of sensors (for cost efficiency);
    \item Capturing heterogeneous spatial locations; and 
    \item Maximizing prediction accuracy
\end{enumerate}

The rest of this paper outlines the background of spatial-temporal data characterization as a foundation for sampling optimization, an overview of the graph model-based methodology for predicting spaces with similar comfort conditions, an overview of the case study data sets, and the results of three sampling scenarios tested to show the theoretical value of occupant feedback sampling based on different spatial zoning selections.

\section{Background of spatial-temporal data characterization}
\label{sec:background}

The foundation of using the BIM-contained digital twin data is based upon the extraction and analysis of spatial-temporal data. This section outlines the literature on forming spatial models and using them for adaptation.

\subsection{Building spatial-temporal data}
\label{sec:st}
Spatial-temporal data (also known as \emph{spatio-temporal} or \emph{spatiotemporal}) records object state, an event, or a position over a time period. It is divided into three distinct types~\cite{Ansari2020SpatiotemporalReview}: \emph{1) events}, \emph{2) geo-referenced time-series data}, and \emph{3) trajectories}~\cite{PelekisLiteratureModels,Boulmakoul2012MOVINGQUERIES}. \emph{Events} refer to a distinct occurrence or action happening in different locations without necessarily being correlated to each other. An example would be the giving of feedback of a building occupant, which is registered in different locations and different times~\cite{Kisilevich2009Spatio-temporalClustering}. \emph{Geo-referenced} data are stationary in specific locations, but the measurements change over time. Environmental sensors (\emph{point}) and space boundaries (\emph{polygon}) are examples of geo-referenced data. \emph{Trajectory} data encompasses both the location and the readings change over time, where the whole histories of the moving object are preserved. Examples of trajectories in the built environment are people moving in buildings, vehicles, and drones. The current research only focuses on indoor spatiotemporal data. Thus, outdoor positioning methods such as the Global Positioning System (GPS) are insufficient because their signals have limited coverage in indoor environments. 


To be able to perform computational analysis on the data, it is necessary to convert continuous data into sample points -- a process called data discretization. For temporal data, this means taking snapshots of the data each time interval or time cycle. For spatial data, there are several methods and techniques based on the application.

\subsection{Spatial discretization}
\label{sec:discretization}

\begin{figure*}[!h]
	\centering
	\includegraphics[height=0.95\textheight]{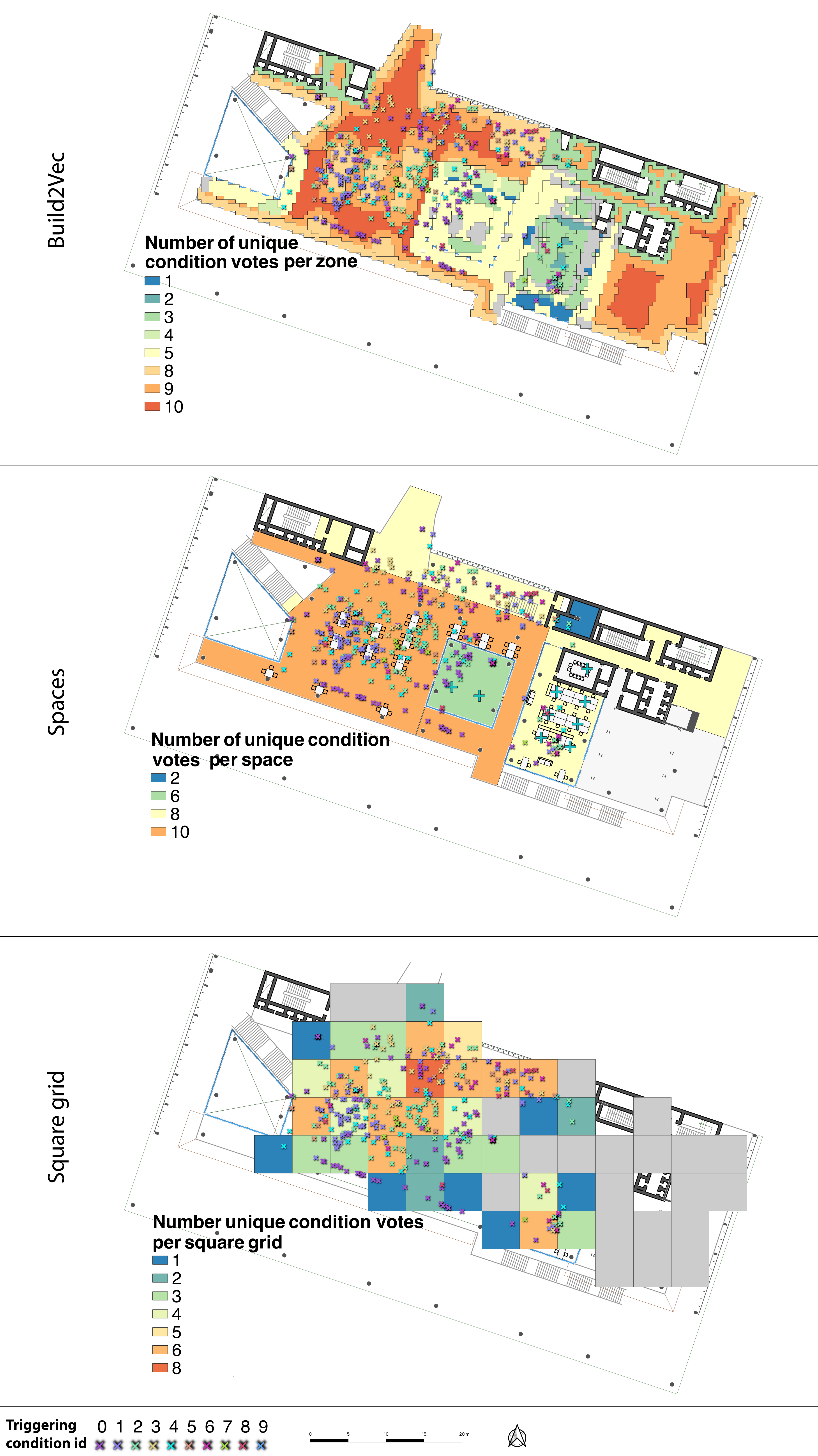}
	\caption{A comparison between three different spatial sampling methods. The color map represents the number of unique condition vote (the x ticks) at each zone.}
	\label{fig:comparison_between_scenarios}
\end{figure*}

Spatial discretization (also known as domain decomposition) is a computational geometry field of study concerned with transferring continuous spaces, models, equations, and variables into their discrete equivalents in the form of finite elements (cells). Each cell share some nodes and edges with its adjacent cells. \textit{Linked lists}, and \textit{graphs} are usually used to store finite element grids efficiently. The scope of the current research only focuses on 2D finite elements, which can be represented as a grid of points or squares. Figure \ref{fig:comparison_between_scenarios} shows the spatial characterization of three scenarios extracted from two occupant preference data sets. The illustrated methods include the division of the floor plate into regular 4x4 meter squares, using the conventional spaces created by the floorplan, and the use of adaptive mesh refinement. The segmentation of the spaces using sampling optimization can be seen, and the color palette for the figure shows the number of unique thermal preference feedback points collected from each zone. The adaptive mesh refinement technique shows the most diverse range of comfort feedback sampling, which implies that there will be the most significant impact in reducing redundant sampling points with targeted feedback triggers.


The terms \textit{meshing} and \textit{mesh refinement} refer to the process by which finite elements are formulated and optimized~\cite{Gander2012}. Several meshing algorithms are found in the literature and are used to develop an accurate and efficient mesh size. The meshing grid's density can be specified to ensure higher accuracy and computational efficiency. Using a fixed density spatial discretization is common in building-related analysis problems (daylight, depth map, and visibility analysis). However, a more efficient solution is to use \emph{adaptive mesh refinement}~\cite{Huang2011}.

\begin{figure}
	\centering
	\includegraphics[width=\linewidth]{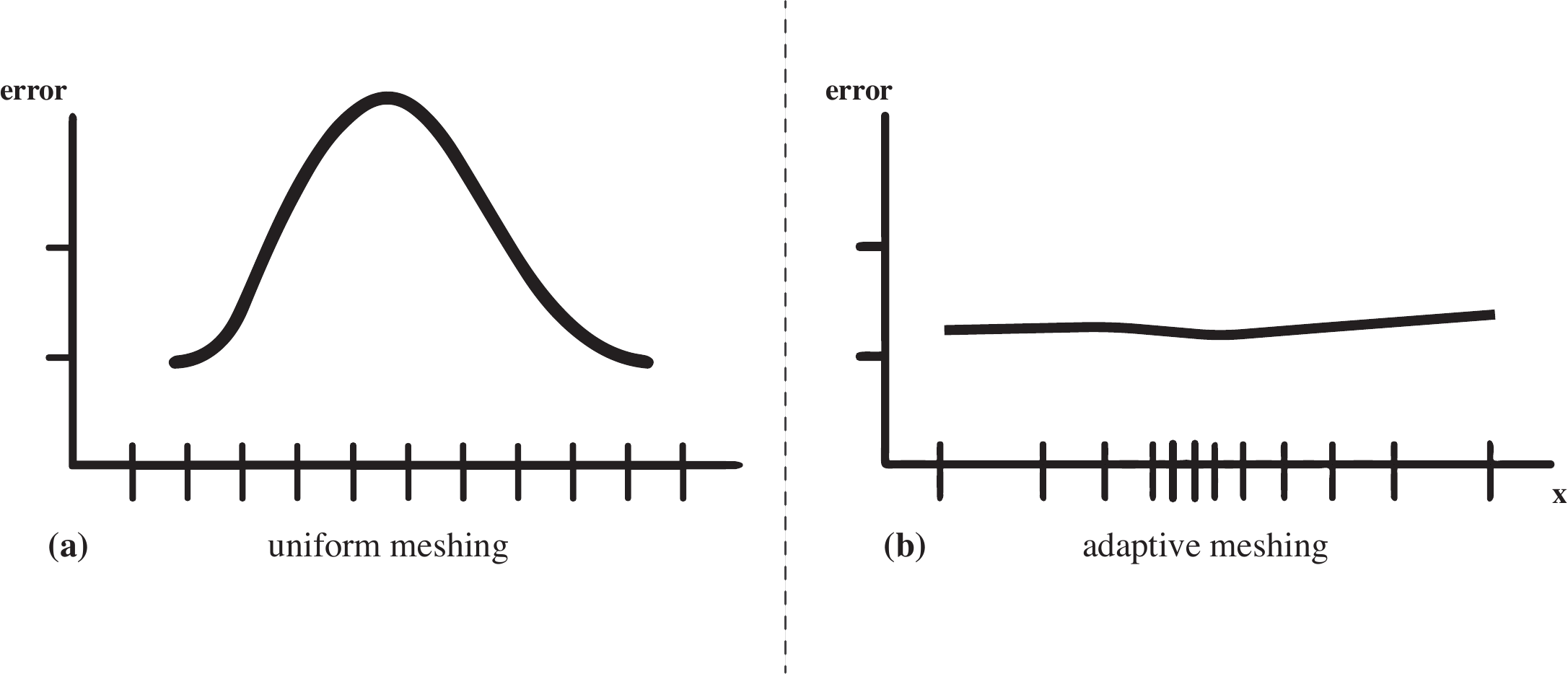}
	\caption{Equidistribution of error algorithms equally distribute the error along the mesh elements (one dimensional FE in this case). This simplified example shows (a) a uniformly distributed meshing, while the error is not equally distributed and (b) adaptive meshing which causes the error to be equally distributed.}
	\label{fig:error_refinement}
\end{figure}

Adaptive meshing is the process of refining or coarsening a meshing grid based on the adjacent mesh elements.Figure \ref{fig:error_refinement} illustrates a simple example of the use of adaptive meshing to distribute error equally. This process comprises three main sub-processes: 1) an optimal-mesh criterion, 2) an error indicator, and 3) an algorithm or strategy to refine and coarsen the mesh. More details on meshing and mesh refinement can be found in previous studies~\cite{Lhner2008AppliedTechniques}.

\begin{figure}
    \centering
    \includegraphics[width=\linewidth]{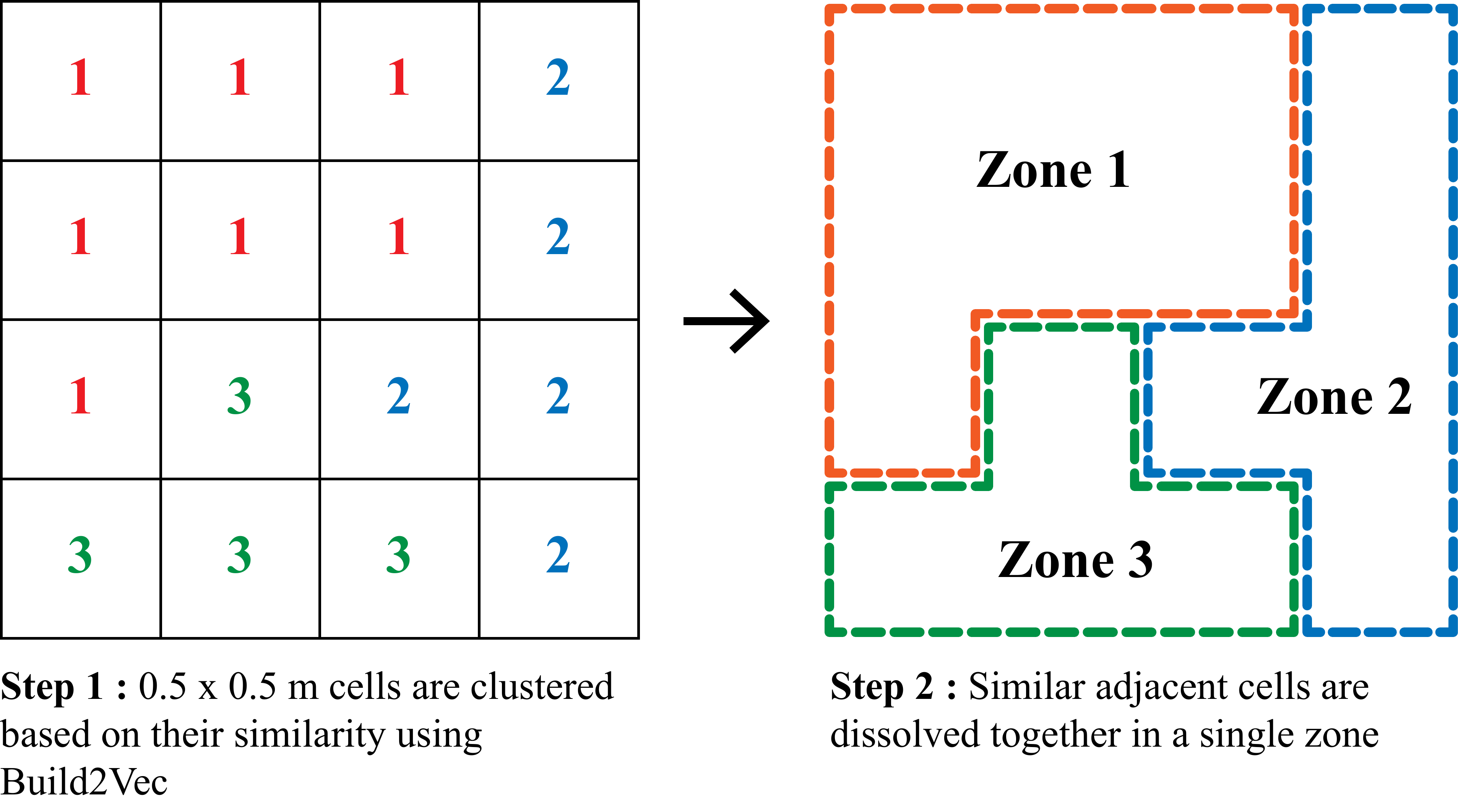}
    \caption{A simplified example of an adaptive mesh refinement of a floor plan.}
    \label{fig:discretization_coarsen_illustration}
\end{figure}

In this research, we use mesh refinement to create regions of spatial similarities. We start by using a fine meshing grid (0.5x0.5 meter), and then this grid is coarsened at each iteration based on the spatial similarity between these grid points. Figure \ref{fig:discretization_coarsen_illustration} outlines a simplified example of this process. This spatial similarity is extracted from the building spatial data (spaces, walls, curtain walls, doors, and stairs) using spatial proximity and graph embeddings (Build2Vec). 


\subsection{Spatial proximity (Spatial nearness)}
\label{sec:spatial_proximity}

\begin{figure}[!h]
	\centering
	\includegraphics[width=\linewidth]{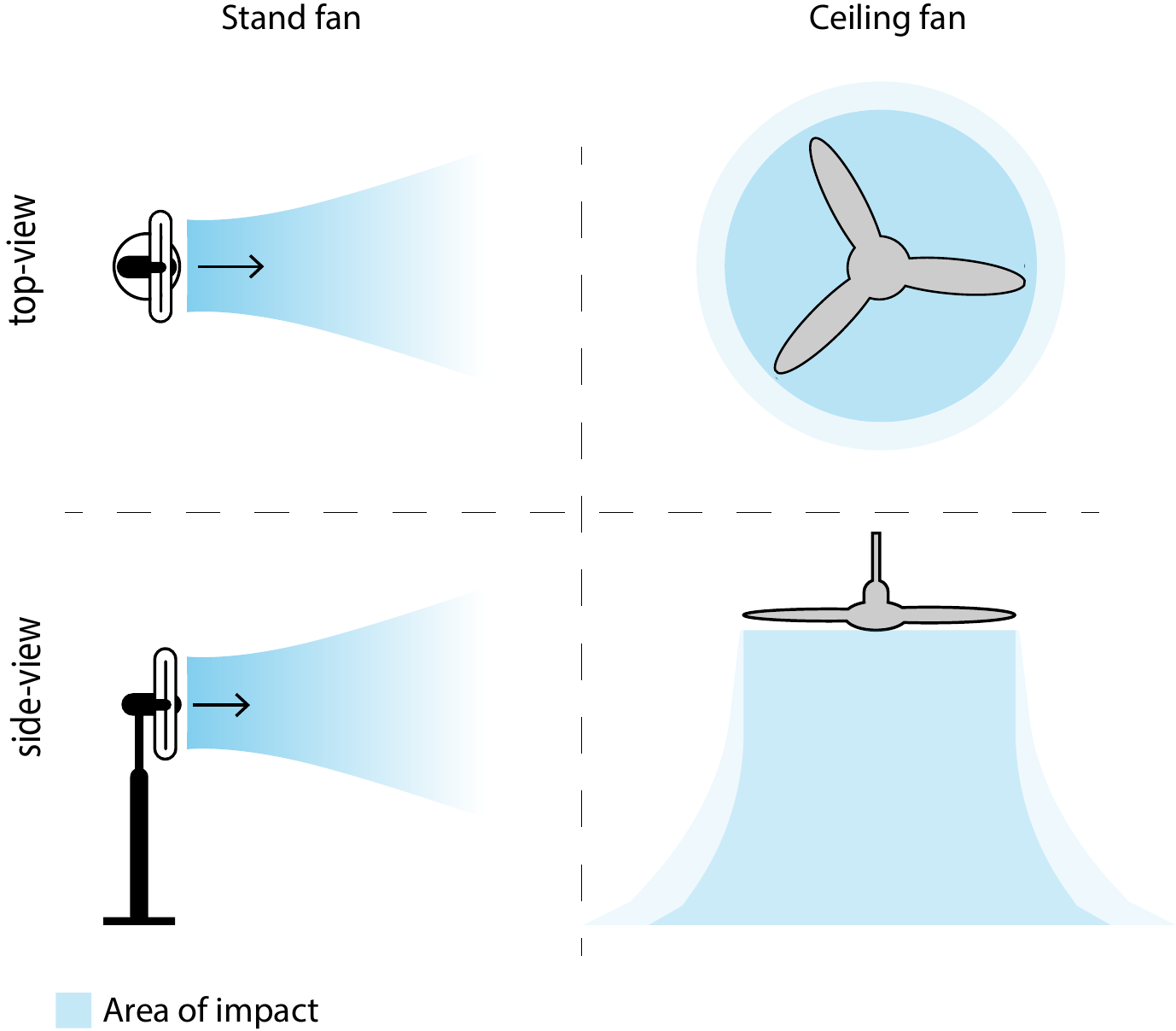}
	\caption{Difference in AoI between a ceiling fan and a standing fan}
	\label{fig:ceiling_stand_fan_aoi}
\end{figure}

In the spatial reasoning field, the components of \emph{spatial proximity} or \emph{spatial nearness} are used to describe how near two objects are to each other. Spatial proximity is an application-oriented concept. For example, \emph{near to the window} can be differently defined based on the application (e.g., daylight, view to outside, glare, noise, air quality, or radiation). Thus, the Area of Impact (AoI) notions are used to describe the buffer area within which each spatial element influences other subjects \cite{kettani1999spatial}. Another critical study introduced the term Impact Area which takes into account the nature of the object and the surrounding environment~\cite{BRENNAN201288}. Impact Area is extended beyond distance proximity to other cognitive, perceptual, and environmental aspects. The current research only focuses on the thermal comfort-related spatial proximity from different spatial objects such as windows, walls (solid and curtain walls), doors (specifically glass doors), and HVAC equipment (ceiling and stand fans, and air conditioning outlets). Figure \ref{fig:ceiling_stand_fan_aoi} shows the difference in AoI between a ceiling fan and a standing fan as an example. 

\subsection{Graph embeddings of building data (Build2Vec)}
\label{graph_embeddings_section}

Graph data mining (or graph data analysis) is the process of gaining knowledge from graphs. Many practical applications depend on graph data analysis techniques such as node classification, node recommendation, and link prediction \cite{Cai2018}. However, most graph analytics methods suffer from a high computational cost. Thus, \emph{graph embeddings} are developed to reduce the computational complexity of graphs by representing graphs in a lower-dimensional space while preserving the original graph's data. This approach helps simplify complex data such as those from the buildings \cite{Cai2018}. 

Buildings' spatial objects are primarily structured in pairwise relations. For example, a building (as an object) has many floors (each floor is an object), each floor has spaces, and each space consists of walls, floors, and ceilings; each wall can have one or more doors or windows. Additionally, each object consists of spatial representations, i.e., the shape and the geometry of the object (which can be represented as a sequence of coordinate points/lines/polygons). Build2Vec is used in this study to extract similarities between different locations in the building by embedding each mesh element into a lower-dimensional vector using Graph Neural Networks (GNN)~\cite{Abdelrahman2017,abdelrahmanbuild2vec}. These vectors are used to cluster the floor plan into different clusters that correspond to their spatial area of influence. 


\section{Methodology}
\begin{figure}[t]
	\centering
	\includegraphics[width=\linewidth]{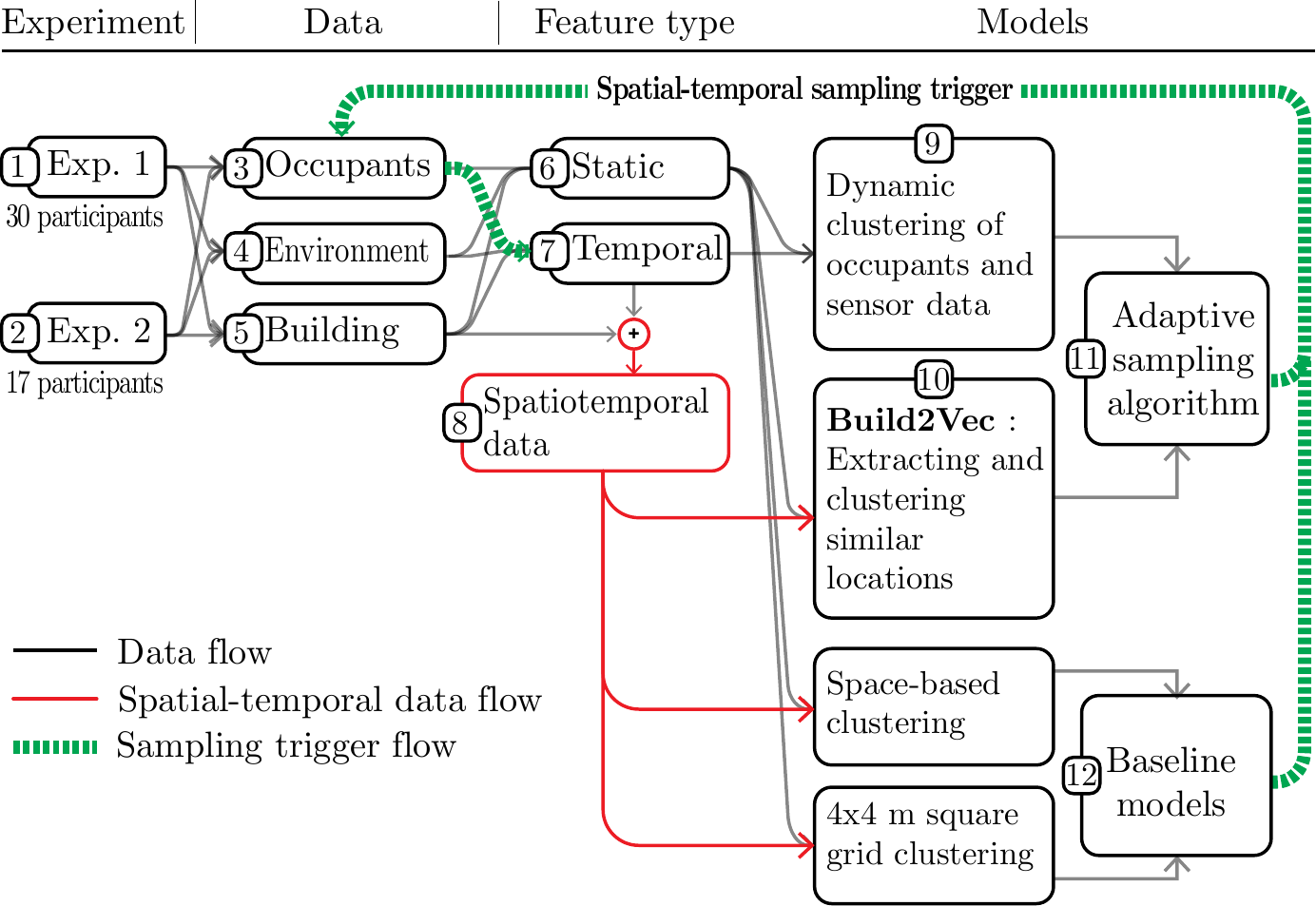}
	\caption{Overview of the implementation on the case study to test the ability of adaptive sampling to improve performance while reducing the amount of data needed. Two studies (1-2) created three data types (3-5) that are used to extract input features (6-8) to be used in a clustering and transformation process (9-10) for a comparative theoretical analysis of the effectiveness of the proposed sampling approach versus several baselines (11-12).}
	\label{fig:work_flow_of_the_methodologyAsset5}
\end{figure}

This section proposes a more balanced data collection and processing method called adaptive spatial-temporal sampling. A summary of the methodology workflow is shown in Figure \ref{fig:work_flow_of_the_methodologyAsset5}. This framework shows two experiments that have been conducted to collect spatiotemporal data from occupants, the building, and the environment. These two sets of thermal comfort data were used for the analysis from previous studies~\cite{buildsysshortpaper2021,Jayathissa2019-kg}. These data sets can be found in open-access Github repositories\footnote{\url{https://github.com/buds-lab/humans-as-a-sensor-for-buildings}}$^,$\footnote{\url{https://github.com/buds-lab/longitudinal-personal-thermal-comfort}}. Both studies were conducted on the campus of the National University of Singapore (NUS). The test setup dictated that participants were to conduct their business in the test-bed building as usual and give feedback during weekday working hours, i.e., from 8:00 AM to 6:00 PM (Monday to Friday). Their feedback focused on the characterization of the thermal preference of the occupants and the spaces towards machine learning-based objectives. A thorough overview of the case study spatial scope is covered in previous work that used vector-based models to predict thermal preference~\cite{Abdelrahman2022-aj}. It should be noted that the sampling methodology for these studies was not optimized and that these data were collected prior to developing the adaptive sampling method. Therefore, this study discusses a set of virtual scenarios based on sampling optimization techniques that could vastly improve the efficacy of data collected in future deployments. 

Three data types are used in this analysis. The first is the occupants' data, including an initial onboarding survey for occupants (static data) and their thermal comfort perception, heart rate, and near body temperature (temporal data). The next type is the environmental data from the case study building, including data from sensors (outdoor temperature data). The last data type is the building spatial data that are collected from the BIM model of the building and are converted into a graph structure. Building spatial data can be static, such as walls and floors, or dynamic, such as opening/closing windows, turning on/off equipment, and moving furniture from one location to another. Keeping track of the building spatial data dynamics constitutes a limitation in the current study. The method includes a step that used occupant and environmental data clustered into ten different triggering conditions. The process also includes the transformation of the building spatial data from the three data types that were dynamically converted into graph embeddings representation using Build2Vec. Then the building is decomposed into 20 clusters of regions that cover all the different combinations of spatial elements. This process is explained in detail in Section \ref{section:triggeringConditions}. The next step includes comparing the adaptive sampling algorithm in which redundant occupant feedback responses are avoided using different spatial contexts. This comparison includes the use of two conventional baselines. The first baseline is the conventional spatial sampling scenario in which occupants are asked to work in specific spaces, and data would be collected within \emph{evenly distributed temporal} intervals. The second baseline is a control scenario in which the building floor plate is subdivided into a normalized 4x4 meter square grid. We use 4x4 squares, particularly as a proxy for an \emph{evenly distributed spatial sampling} method. 

\begin{figure*}[!h]
	\centering
	\includegraphics[width=0.9\linewidth]{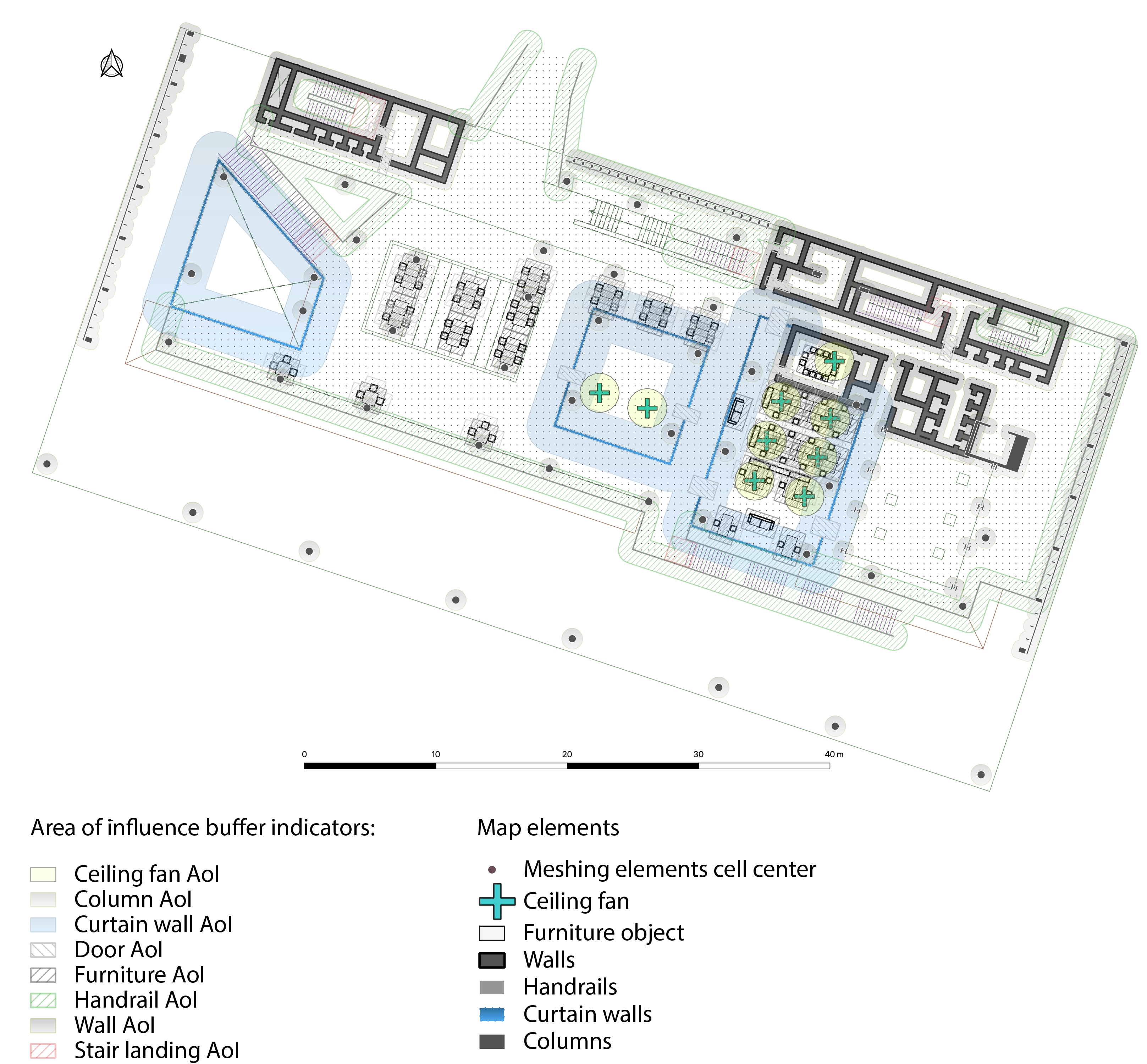}
	\caption{Spatial data pre-processing in which the spatial data are extracted and projected to the EPSG:4326 system. The spaces were then discretized into finite elements. The area of impact buffer zone around each spatial element is illustrated.}
	\label{fig:spatial_data_preprocessing}
\end{figure*}

A detailed overview of the three types of data that were collected in these studies and used in this analysis is as follows:
\begin{enumerate}
	\item \textbf{Occupants' data} - Personal subjective and physiological data were collected including:
	      \begin{itemize} 
	      	\item Participants were asked to fill in an on-boarding survey. The survey contains physio-psychological questions such as gender, height, weight, and Big-Five personality traits~\cite{buildsysshortpaper2021}.
	      	\item Occupants were asked to wear a smartwatch, move freely, and answer thermal comfort questionnaires during their normal work time in the building using the Cozie smartwatch platform\footnote{\url{https://github.com/cozie-app/}}~\cite{Jayathissa2019-kg}. Each participant was asked to give 80 feedback responses with minimum 15 minutes intervals between each response. 
	      	\item The heart rate, near-body temperature, and skin temperature data of each participant were also collected.
	      	\item Occupants' location was collected in the building using an indoor localization system with accuracy ranges from 0.25 to 3 meters. 
	      \end{itemize}
	\item \textbf{Environmental data} - Environmental data were collected from various IoT sensor sources such as:
	      \begin{itemize}
	      	\item Three weather stations were installed in the outdoor spaces to collect the outdoor weather data, including (temperature, humidity, and CO2). 
	      	\item Low-cost sensing sensors were installed in each space and placed in locations that are likely to represent the space and where there is a power plug. However, these data have been excluded since the focus of this paper is spatial heterogeneity.
	      \end{itemize}
	\item \textbf{Building Spatial data} - The spatial data were extracted from the BIM model and preprocessed as follows:
	      \begin{itemize}
	      	\item Spatial data were extracted from spaces, furniture, HVAC elements, doors, curtain walls, stair landings, and windows.
	      	\item Spatial projection was used to transform the data from the metric coordinate system to the GPS coordinate system EPSG:4326.
	      	\item Spatial discretization was used to subdivide the floor plan into a 0.5 x 0.5 m meshing grid.
	      	\item Spatial proximity was extracted for each spatial element by adding a buffer area that defines the area of impact of each spatial element. Figure \ref{fig:spatial_data_preprocessing} shows an example of this process.
	      	\item The graph representation of the building data was formed from the relation between each spatial element and the nodes that fall into its proximity using a graph database management system calledj neo4J\footnote{\url{https://neo4j.com/}}.
	      \end{itemize}
\end{enumerate}

\subsection{Build2Vec transformation}
\label{build2VecSection}
\begin{figure*}[!h]
	\centering
	\includegraphics[width=\linewidth]{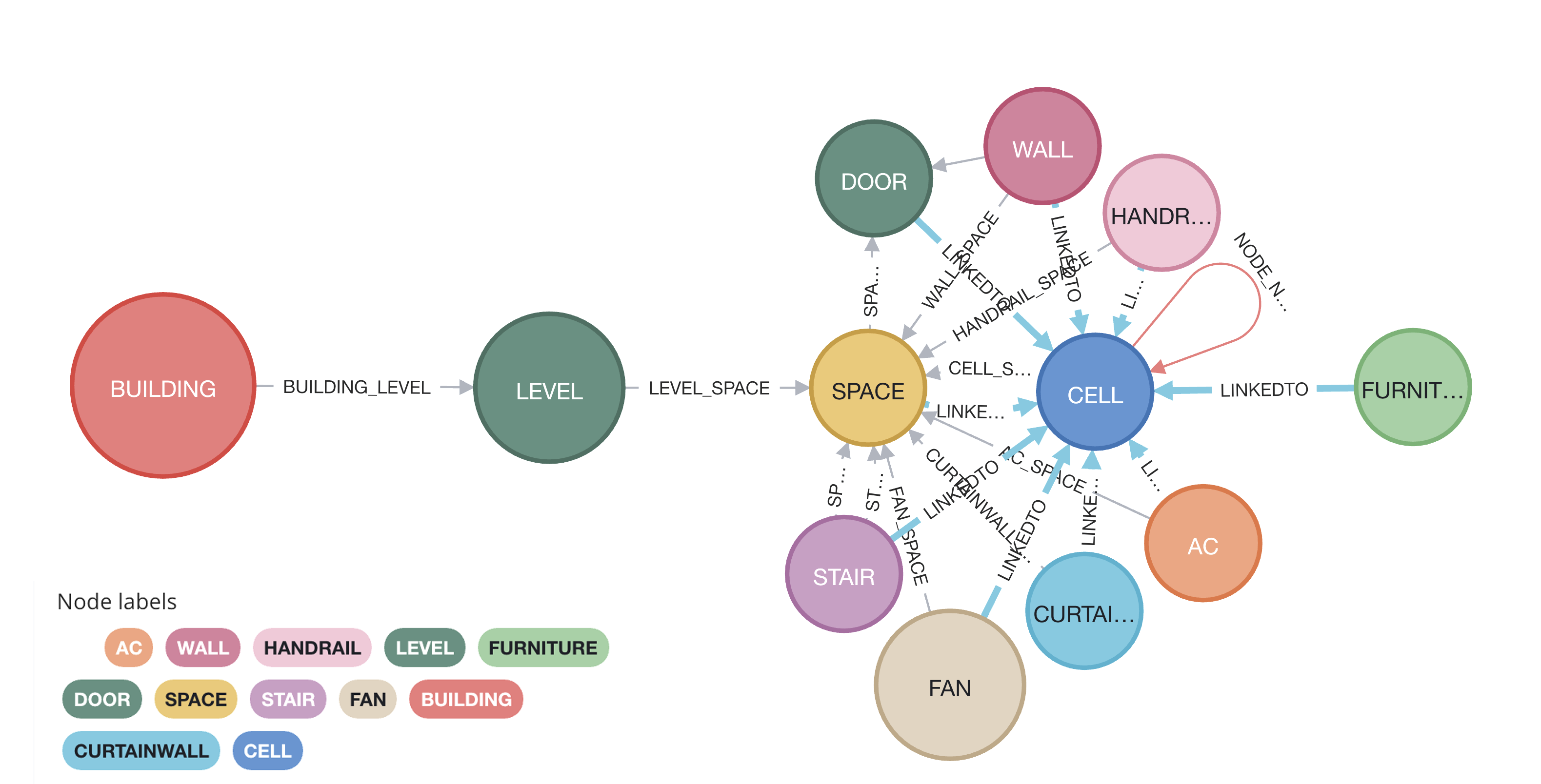}
	\caption{A schematic graph structure representation of the building spatial data}
	\label{fig:Figure_10_graph_schema}
\end{figure*}

Figure \ref{fig:Figure_10_graph_schema} illustrates how Build2Vec is used to convert the graph structure into lower dimension vector representation where the graph data are embedded \cite{Abdelrahman2017}. The embedding vector, in this case, represents each spatial object and the proximity information. Cosine similarity was used to calculate the similarity of spatial objects' vectors (Equation \ref{eqn:sim}), where $A_{i}$ and $B_{i}$ are components of vector $A$ and $B$ respectively.

\begin{equation}
	\text { similarity }=\cos (\theta)=\frac{\mathbf{A} \cdot \mathbf{B}}{\|\mathbf{A}\|\|\mathbf{B}\|}=\frac{\sum_{i=1}^{n} A_{i} B_{i}}{\sqrt{\sum_{i=1}^{n} A_{i}^{2}} \sqrt{\sum_{i=1}^{n} B_{i}^{2}}}
	\label{eqn:sim}
\end{equation}

\begin{figure*}[!h]
	\centering
	\includegraphics[width=\linewidth]{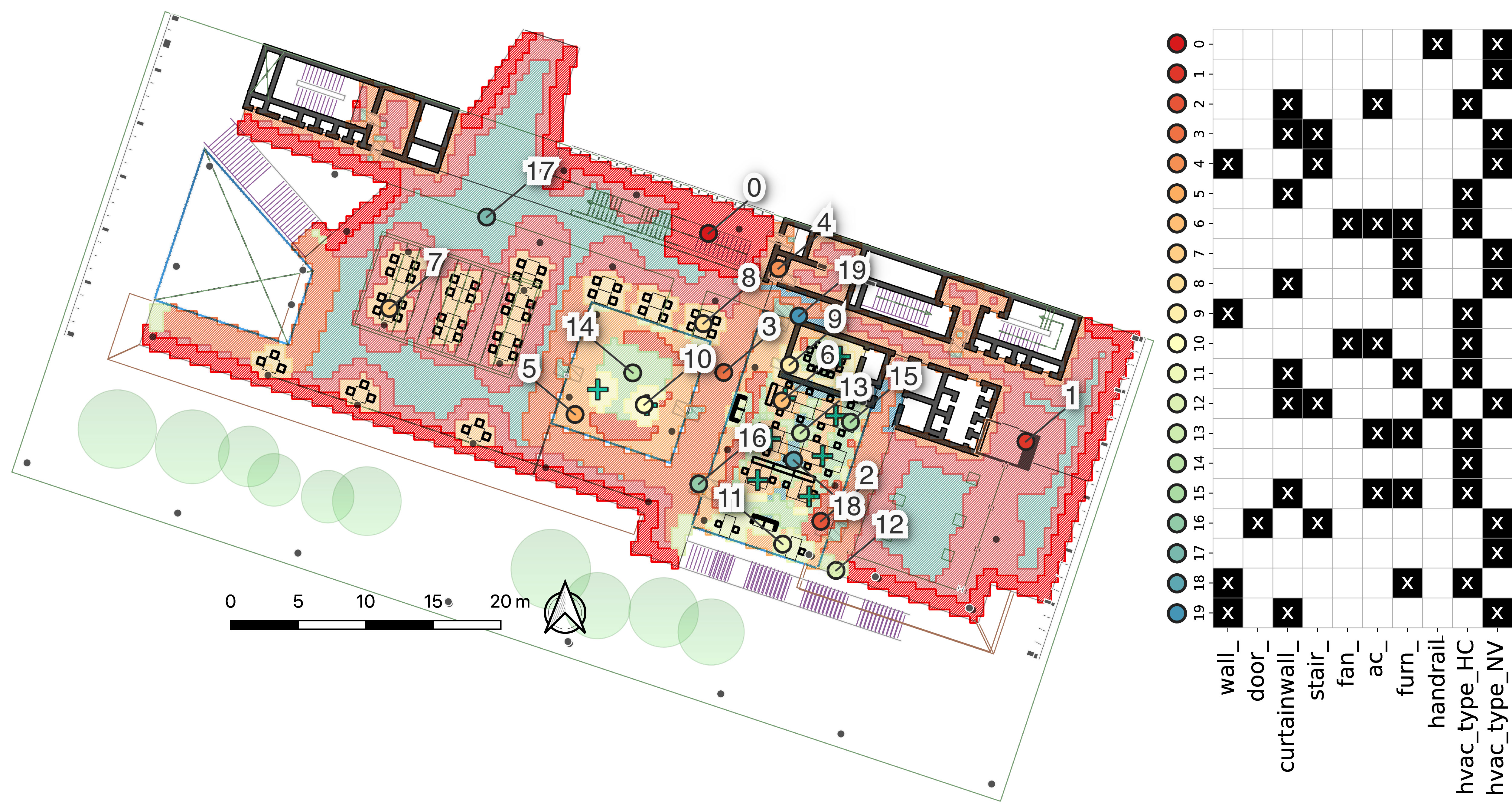}
	\caption{The Build2Vec spatial proximity clustering result. Each cluster has unique spatial properties to capture the heterogeneity of spaces. It is noted that Cluster 1 and 17 both have the same properties. However, Cluster 17 represent locations that are far from any other spatial objects while Cluster 1 is closer to other spatial objects}
	\label{fig:spatial_heterogenous_zones}
\end{figure*}

The Build2Vec similarity vectors were used to cluster the cells into 20 groups. The number of clusters was selected to represent a large variety of spatial conditions. However, the selection of a fixed number that may not be optimal is a limitation of the current study that needs to be addressed in future implementations. The next step is coarsening the mesh grid such that each group of similar cells is dissolved into one element. The mesh coarsening process resulted in 20 different geofenced polygons. These polygons serve as the new spatial zones that represent the heterogeneity of the floor plate. Figure \ref{fig:spatial_heterogenous_zones} shows the 20 zones and their corresponding spatial proximity features. The zones are coded into 20 unique identification numbers (ids) from 0 to 19. 

\subsection{The triggering condition}
\label{section:triggeringConditions}
\begin{figure}
    \centering
    \includegraphics[width=\linewidth]{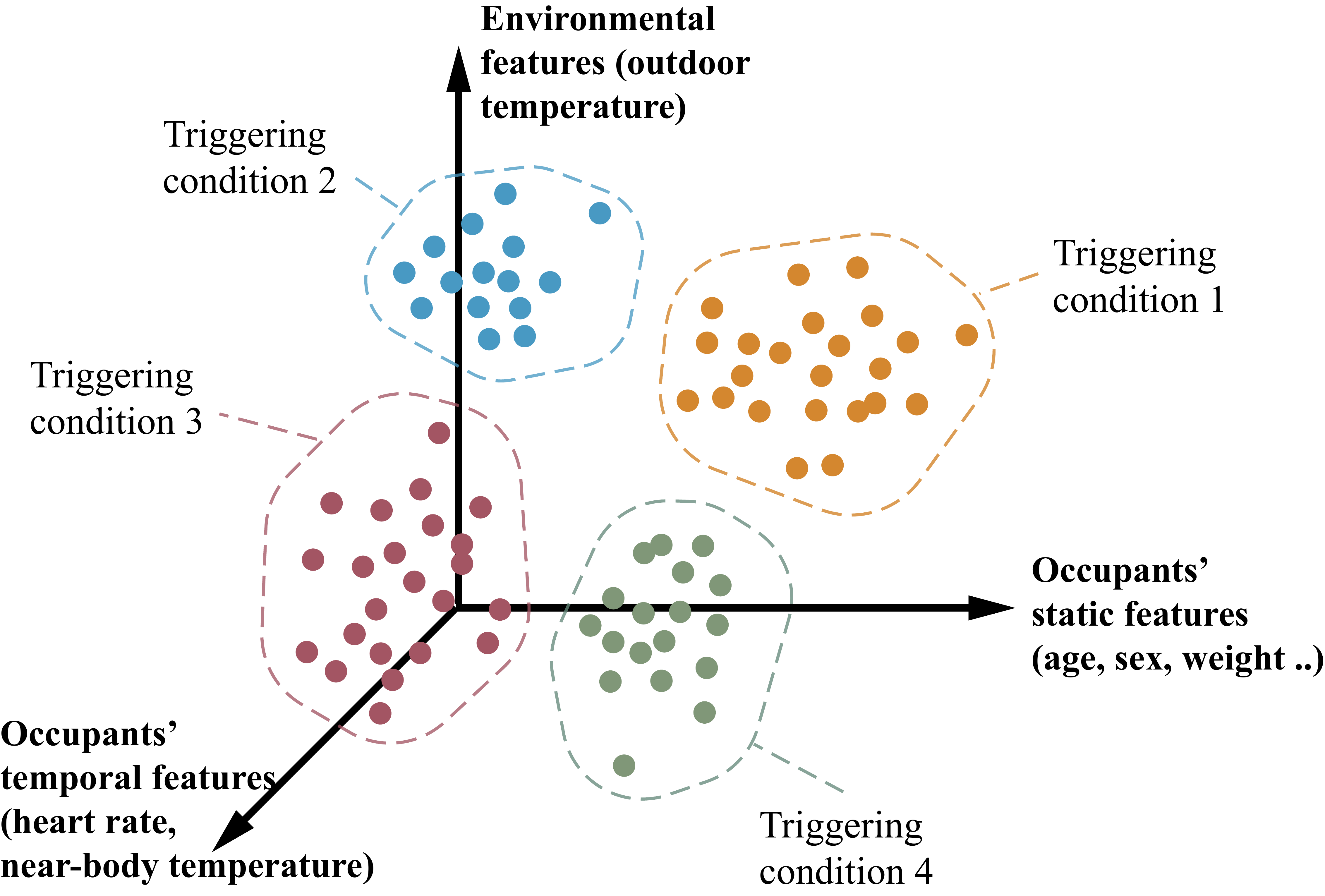}
    \caption{Illustration of clustering the triggering conditions into different clusters}
    \label{fig:triggeringConditionsAsset}
\end{figure}

\begin{figure*}[!h]
	\centering
	\includegraphics[width=\linewidth]{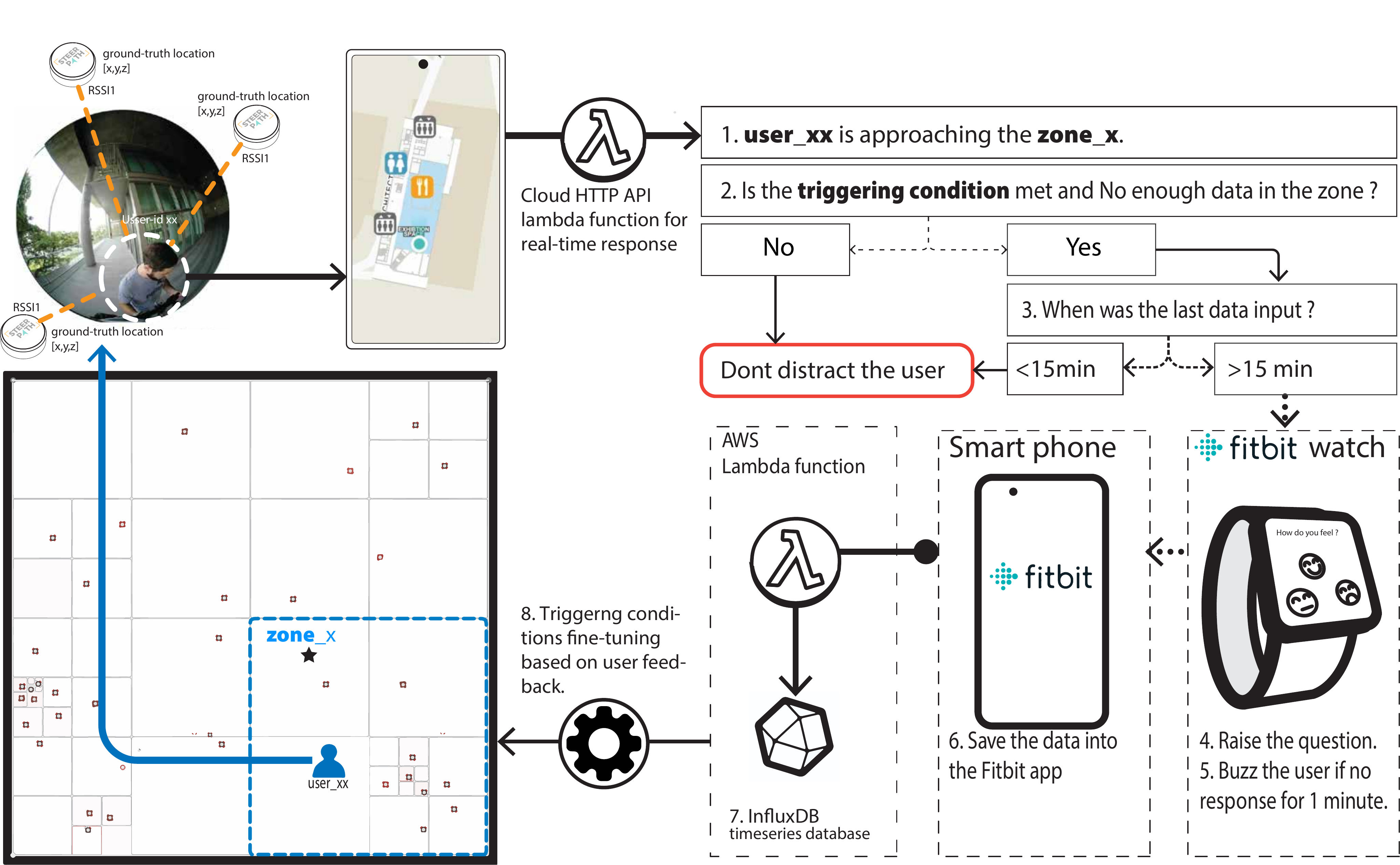}
	\caption{The theoretical triggering conditions framework: 1) The user location within spatial zones is detected using real-time indoor localization and spatial reasoning, e.g. (\texttt{POINT WITHIN POLYGON}), 2) If there is inadequate data in this zone that meet the current condition, the algorithm check for 3) the last time when the use gave feedback if the latest input was more than 15 minutes, 4-7) the user is prompted to give feedback  8) Finally, the triggering conditions are fine-tuned based on the new user response feedback.}
	\label{fig:adaptive_spatial_temporal_sampling}
\end{figure*}

In this study, the key focus is to create a theoretical simulation of the benefits of strategically targeting feedback by creating triggering conditions for prompting the participants to give subjective feedback. Past studies have used fixed-time intervals, scheduled and manual triggers, and environmental conditions to collect feedback in an optimal way~\cite{DuarteRoa2020TargetedConditions}. The proposed method in this study builds upon that work by using the spatial context as a factor in the triggering process. The first step in this process is clustering occupants and environmental data into ten different groups that represent the triggering conditions. Categorical data from occupants are then obtained from the onboarding survey. A preprocessing step is conducted to select the two most unique crucial data types and their representative temporal data. We found that outdoor temperature correlates well with other data and, therefore, can be used as a proxy for other data points. At the same time, the heart rate feature exhibits the lowest correlation with other data, which means that it is the most distinct data point. Thus, we chose the outdoor temperature and the heart rate as the two representative temporal features of the current dataset. We used K-means clustering to cluster the personal and temporal data into 10 clusters representing the triggering conditions (Figure \ref{fig:triggeringConditionsAsset}). For example, whenever the outdoor temperature, the heart rate, and the personality type match one of the clusters, the user gets prompted to complete a survey. Figure \ref{fig:adaptive_spatial_temporal_sampling} illustrates an overview of the theoretical framework for adaptive sampling tested in this work. 

\subsection{Sampling quality metrics}
Four metrics are used to evaluate the quality of the sampling process. These metrics are designed to be applied to scenarios to compare the (1) \emph{scalability}, (2) \emph{adequacy}, (3) \emph{redundancy}, and (4) \emph{reliability} or spatial characterization options for targeting samples.

\subsubsection{Scalability}
The scalability of a system is defined as its ability to handle a growing amount of load by adding more resources. Scalability is an essential factor for generalizing indoor environmental quality studies on a large scale. In this research, we use three resource parameters to measure the sampling scalability: 

\begin{enumerate}
    \item Space scalability ($S_s$): How complex will the system become by adding more spaces. That is, how many zones need to be covered by the survey experiment given the total number of space $S_{st }$. Table \ref{tab:hvac_systems_at_sde42} shows the total number of spaces at each floor of the case study. 
    
    \item Floor area scalability($S_a$): How many spaces/zones are required to cover the total floor area of the building $S_{at}$. $S_a$ is a normalized number of spaces per square meter metric. It is obtained from $(1-\frac{N}{A})$ where $N$ is the number of zones, and $A$ is the gross floor area. 
    
    \item Occupants' scalability ($S_o$): This is an indicator of the sample size required to conduct a survey for the total number of population $S_{ot}$. 
 
\end{enumerate}

\begin{table}[htbp]
  \centering
  \caption{Summary of the HVAC system types and zones in the case study building.}
  \label{tab:hvac_systems_at_sde42}
  \includegraphics[width=\linewidth]{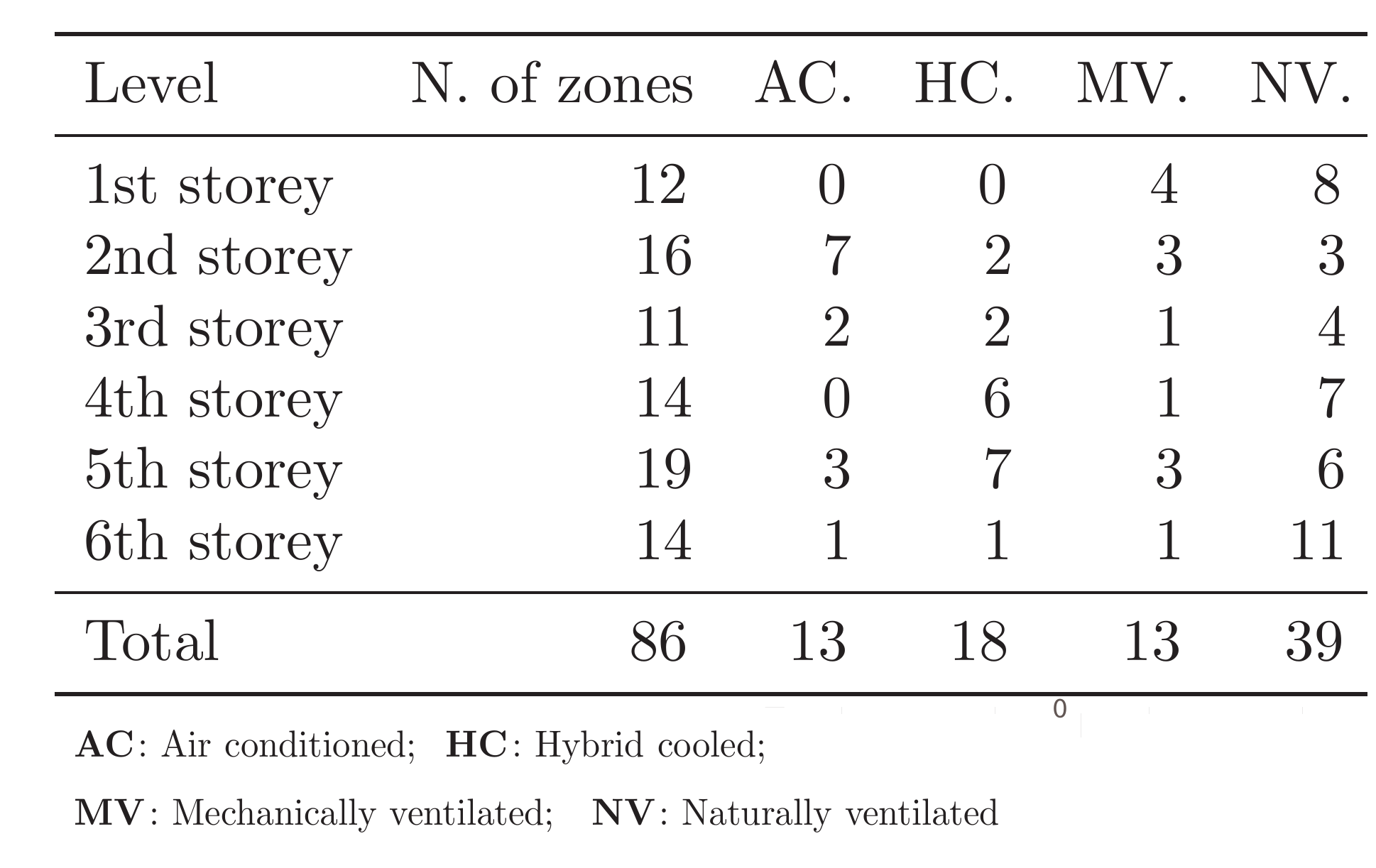}
\end{table}%

Equations \ref{eq:cochran_1} and \ref{eq:cochran_2} show how the sample size required for IEQ surveys is calculated~\cite{Techniques,Cochran1977}:

\begin{equation}
n_0 = \frac{Z^2 p q}{e^2}
\label{eq:cochran_1}
\end{equation}

In Equation \ref{eq:cochran_1}, $n_0$ is the pooled sample size, $Z^2$ is the area under the normal curve with 0.05 level of significance, $p$ is the estimated proportion, $q = 1-p$, and $e$ is the error margin. Next, the actual sample size is calculated using Equation \ref{eq:cochran_2} from the pooled sample size $n_0$. 

\begin{equation}
n = \frac{n_0}{1+\frac{(n_0-1)}{N}}
\label{eq:cochran_2}
\end{equation}

In Equation \ref{eq:cochran_2}, $n$ is the sample size and $n_0$ is the pooled sample size obtained from Equation \ref{eq:cochran_1}. For this study, the population size is 650 occupants, with an error margin of 11\%, confidence level of 90\%, and response distribution of 50\%; the total sample size needed is 50. 

The overall scalability quality $S$ is calculated using Equation \ref{eq:eq3}.

\begin{equation}
S=  \frac{1}{3} \sum_{i \in \{s, a, o\}} \sigma\left(\frac{S_i}{S_{i\max }-S_{i\min }}\right)
\label{eq:eq3}
\end{equation}

The $\sigma$ in Equation \ref{eq:eq3} is a softmax normalization function obtained using Equation \ref{eq:eq4}

\begin{equation}
\operatorname{\sigma}=\frac{\sum e^{z_{j}}}{\sum_{k=1}^{K} e^{z_{k}}}
\label{eq:eq4}
\end{equation}

The softmax function is used for normalization because it is easy to back-propagate for optimization in real-time scenarios. By taking the derivative of the softmax, the sampling quality could be added as a layer in the neural-network problem (e.g., reinforcement learning).

\subsubsection{Data adequacy and redundancy}
\begin{figure}[!h]
    \centering
    \includegraphics[width=\linewidth]{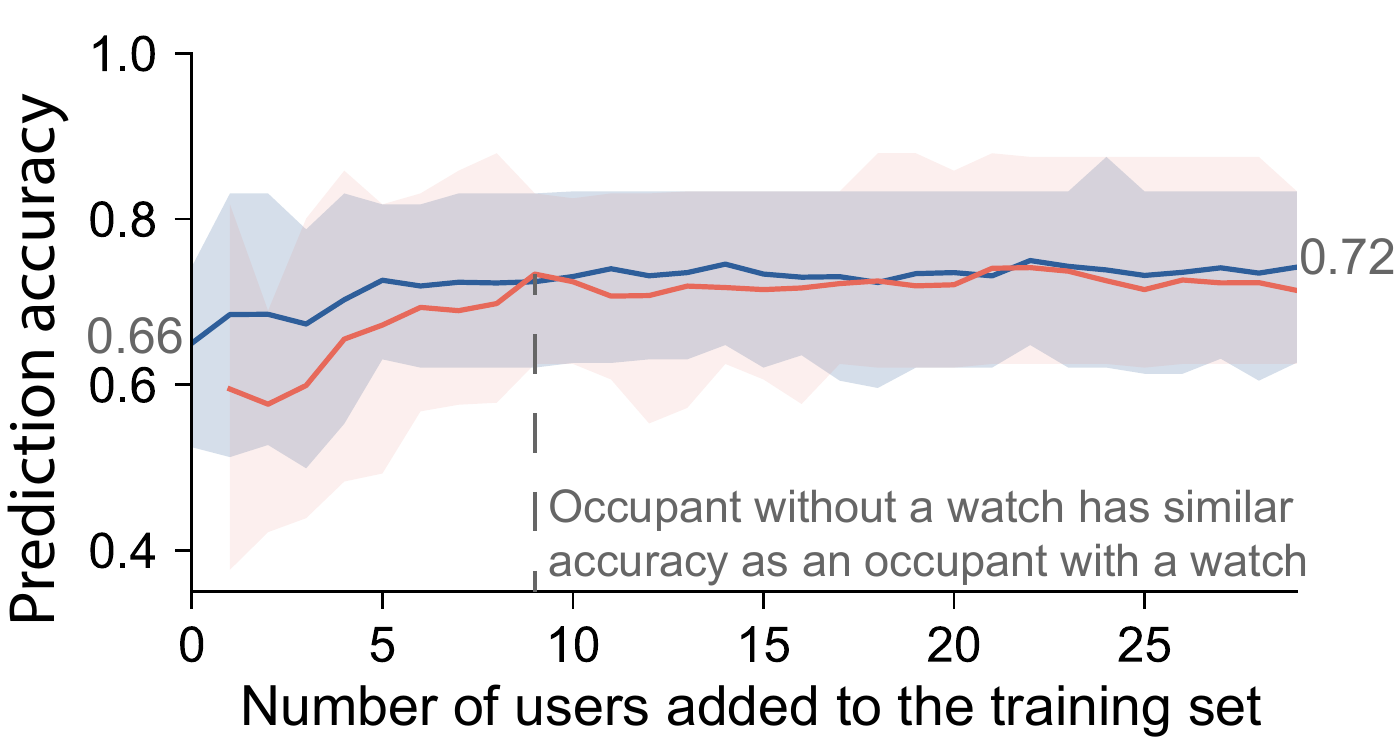}
    \caption{Results from previous work showing that adding more data from users doesn't necessary improve prediction accuracy (adapted from~\cite{Jayathissa2020-pv}).}
    \label{fig:added_data_not_necessary}
\end{figure}

Two metrics are used to indicate the optimal number of votes required. These components are 1) data adequacy and 2) data redundancy. Data adequacy refers to the actual number of unique met-condition votes to the required number of unique votes per spatial zone. In the current research, the required unique votes value is 10 (i.e., at least one vote for each met condition. Data redundancy refers to the extent to which the actual number of votes per condition exceeds the required. Figure \ref{fig:added_data_not_necessary} shows an example of how the prediction accuracy can hit a plateau after adding a certain number of users' data from a previous study~\cite{Jayathissa2020-pv}. We calculate the adequacy/redundancy sampling quality as $Q_s$ using Equation \ref{eq:eq5}.

\begin{equation}
Q_s = \frac{D_u}{x} - \frac{\sum_{u=1}^{x} \left |  {A}_u  -y \right |}{n}
\label{eq:eq5}
\end{equation}

In Equation \ref{eq:eq5}, $D_u$ is the ground truth unique data per zone and $x$ is the required unique data per zone. We also refer to $x$ as the conditions per zone and this value is set to 10 in this study. $A_u$ is the actual ground truth number of votes for each condition, $u$ is the triggering condition, $y$ is the required unique votes for each condition, and $n$ is the total number of votes at each zone. In this study, we use one vote per condition for $y$.

\subsubsection{Reliability}
Reliability refers to the prediction accuracy obtained from applying a sampling method. In this study, we use a baseline model applied to the case study and published in previous work~\cite{Jayathissa2020-pv}. We compare this approach with a different method that uses Build2Vec for thermal comfort prediction~\cite{Abdelrahman2022-aj}.

\section{Results}
\label{sec:results}

This section outlines the results of testing three scenarios of different methods of sampling thermal preference feedback data from occupants. These scenarios utilize the collected data from the two source datasets and implement theoretical processes differentiated by how the context of occupant preference sampling optimization is implemented. The first scenario uses space-based segmentation in which the physical zones of the building's floor plan are used as the spatial optimization context. This scenario serves as a baseline in which a certain amount of feedback from each zone is expected. The sampling process would seek to optimize for this objective. The second scenario is a generic 4x4 meter square uniform grid that can be used as an evenly distributed spatial resolution baseline. The third scenario is the proposed sampling optimization method using the Build2Vec based spatial clustering model. This scenario utilizes the BIM to vector conversion and occupant response similarity extraction as the foundation for creating zones that can be considered consistent. 

The result of the Build2Vec model is a 50-dimensional vector for each cell that was created in this implementation. The Build2Vec model is used to convert the graph into a machine learning-friendly input (embedding matrix). To do this, we use a method to sample from the adjacent nodes of each source node in the graph. However, this process could be computationally expensive. Thus, we use a normally distributed \textit{random walks} for sampling the data from the adjacency nodes. The random walk number is the number of the adjacent nodes from which to sample. The \textit{number of walks} per iteration indicates how deep each walk should go before it stops sampling. After sampling the nodes, we use \textit{window size} to predict a central node based on the surrounding nodes from both directions. This method is inspired by Natural Language Processing (word2Vec), where the central word is predicted from the context within which it falls. Based on best practices, these parameters (random walks, number of walks, and window size) are fine-tuned. In the current study, we used a random walk length of 50, the number of walks was 50, and the sliding window size was 30. These vectors were used to cluster the cell nodes into 20 different spatial zones, as mentioned in Figure \ref{fig:spatial_heterogenous_zones}.

\subsection{Comparison of sampling scenarios}

\begin{figure}[!h]
    \centering
    \includegraphics[width=\linewidth]{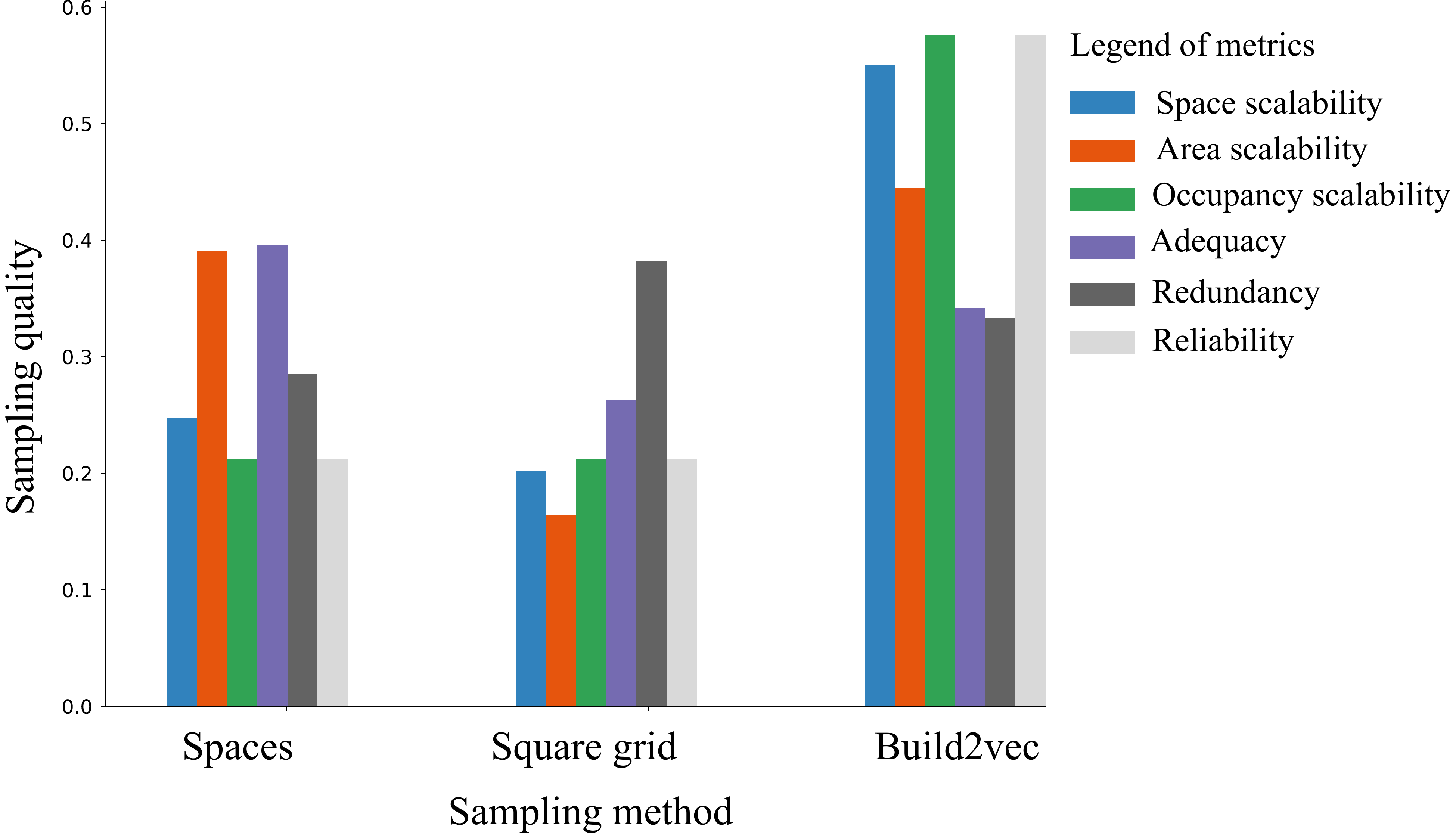}
    \caption{A comparison between different quality metrics of the three sampling methods.}
    \label{fig:quality_metrics_visualization}
\end{figure}

 The result of calculating the four sampling quality metrics is shown in Figure \ref{fig:quality_metrics_visualization}. The Build2Vec method offers the highest overall quality in all metrics except for redundancy, in which it is surpassed by the space-based sampling method. The square grid method shows the worst quality among the techniques. On the other hand, the space-based sampling method shows the highest adequacy quality. This situation is attributed to the relatively large sizes and few numbers of spaces. Only Zone 17 has fulfilled all the conditions attributed to the large floor area of this zone and its location. Whereas other areas such as Zones 2, 6, and 11 are located in closed office spaces not accessible to all participants.

\begin{figure}[!h]
    \centering
    \includegraphics[width=\linewidth]{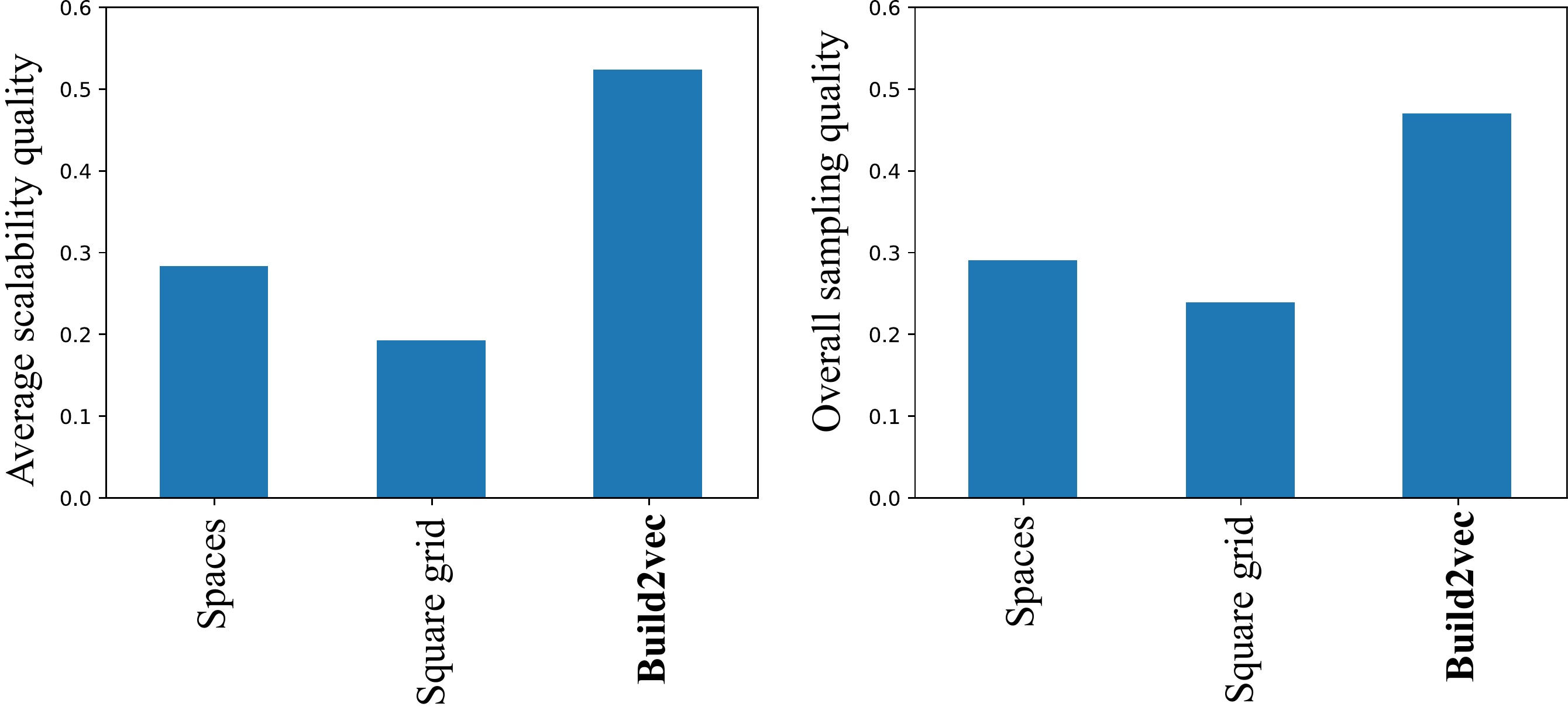}
    \caption{Comparison of the scalability (left) and overall sampling quality (right) metrics across the scenarios.}
    \label{fig:overall_sampling_quality}
\end{figure}

Figure \ref{fig:overall_sampling_quality} shows how the Build2Vec method has higher scalability quality and overall quality compared to the other scenarios. The number of spatial zones is constant, with more spaces/floor area/occupants added to the building. The reason for the constant value is due to the clustering of spatial zones based on their similarity and the clustering of occupants based on their individual personality features. This scenario makes the complexity of this system more efficient (i.e., $O(1)$ constant complexity) than the other two methods, which scale linearly with any added new space, area, or occupant (i.e., $O(N)$ linear complexity). 

Table \ref{tbl:myLboro} outlines the comparison of the benefit of using strategic sampling in terms of the reduction of occupant feedback data points. This table shows the adequacy/redundancy quality for each spatial unit of the three scenarios. The adequacy/redundancy quality is shown both before and after manually removing the redundant votes (those are feedback points that have the same triggering condition within the same zone). Although the \emph{Build2Vec} method shows less adequacy quality than the baseline \emph{Spaces} method, the result from the \emph{Spaces} scenario is prone to decrease by scaling up the experiment. The \emph{Build2Vec} model has a constant scalability complexity (i.e., the spatial clusters will remain 20 $O(1)$). The \emph{spaces}-based model has a linear scalability complexity (i.e., any new spaces will require adding new votes to the experiment $O(N)$). 

\begin{table*}[h]
	\begin{center}
		\caption{Comparison of the three scenarios focused on showing the benefit of strategic sampling of comfort preference information from occupants}
		\begin{tabular}{ p{0.2\textwidth}|p{0.2\textwidth}|p{0.2\textwidth}|p{0.2\textwidth}  }
			\toprule
			{\textbf{Adequacy/ redundancy quality}}& Spaces  & Square grid  & \textbf{Build2Vec}  \\ 
			\toprule
			\footnotesize{Before removing redundant votes} 
            & 
            \footnotesize{Mean 0.38 \newline Median  0.46}
            & 
            \footnotesize {Mean  0.26 \newline Median  0.26}
            &
            \footnotesize{\textbf{Mean 0.47 \newline Median 0.47}}
            \\ \midrule
            \footnotesize{After removing redundant votes}
            & 
            \footnotesize{\textbf{Mean 0.64 \newline Median 0.8}}
            & 
            \footnotesize{Mean 0.23 \newline Median 0.15}
            &
            \footnotesize{Mean 0.53 \newline Median 0.5}
            \\ \midrule
            \footnotesize{Scalability complexity}
            
            & 
            \footnotesize{Linear $O(N)$}
            & 
            \footnotesize{Linear $O(N)$}
            &
            \footnotesize{\textbf{Constant $O(1)$}}
            \\ \bottomrule
        \end{tabular}
        \label{tbl:myLboro}
    \end{center}
\end{table*}

\section{Discussion}
The development of methods to capture human subjective feedback in a scalable and ubiquitous way will enhance several aspects of the built environment and occupant behavior characterization. This section outlines several potential areas of impact from this type of work.

\subsection{Targeting occupant feedback to reduce survey fatigue}
The process of collecting intensive longitudinal data from building occupants in a field setting is an emerging area of research and development that seeks to improve the characterization of the built environment context. Work in this area could further enhance the innovation in adaptive and personal comfort models~\cite{Parkinson2020}. The use of mobile wearable devices enhances the ability to collect such data, but there are still hesitations on the part of practitioners to deploy these methodologies for practical use. This hesitation is due to occupants' reliance on using an application or device to create the data actively. There could be situations in practice where an occupant is being asked to give feedback at a specific frequency as they go about their day, and in return, a reward such as a monetary reward is given for their efforts. The work in this paper increases the probability that the data generated through such an effort would be more meaningful for the purposes of characterization and model training. The use of a strategic sampling of data could reduce the redundancy and prompt the user to give feedback in less frequently encountered situations both temporally and spatially. 

\subsection{Expansion of targeted feedback to noise, privacy, and other occupant comfort objectives}
While thermal comfort is a primary focus of work in occupant-centric building performance, there are numerous other potential areas of indoor environmental quality that could rely on the use of targeted sampling to better characterize spaces with less survey fatigue. The methodology could be used for noise, privacy, lighting, and infectious disease-focused data collection objectives.

\subsection{Limitations and future studies}
This study has several limitations related to the case study context and the breadth of data used in the modeling process. The generalisability of these results is subject to certain limitations. On the one hand, installing a relatively accurate indoor positioning system still requires collecting users' location data in buildings. On the other hand, the objective of this study was to be conducted in a real-time scenario. However, the data used in this study were processed offline, resulting in many redundant data. Reducing the redundant data can be achieved in real-time by two means: 1) ignoring any redundant triggering condition if there are already enough data points at a specific zone, and 2) engaging a real-time bi-directional request-response system that prompts the user whenever a condition is met in a vacant zone. Nonetheless, there will still be zones not fulfilled with data by the end of the experiment. Future research can be conducted to allocate the users (e.g., through a recommendation system) to work from zones where votes are required for the current condition. Another solution is to pre-define these zones prior to the start of the experiment.

As in any field-based study, there are limitations related to the control of variables that are usually managed in a lab-based setting. For example, ensuring that the test participant is in a steady-state comfort state when they leave feedback can be maximized but not fully optimized. Further work in detecting these situations using physiological sensing or additional subjective questions is being explored for future deployments. Another limitation is the fact that 20 spatial zones were chosen an arbitrary parameter for the proposed method. Future studies should consider conducting a spatial-feature importance study to select the number of zones. Furthermore, there are some practical limitations of deploying similar studies in real-life scenarios. For example, providing everyone with a smartwatch and asking them to give feedback regularly is a challenge. Also, the privacy of the participants being tracked in the building is an issue for such experiments. 

\section{Conclusion}
\label{sec:conclusion}
This paper outlines the use of occupant comfort sampling optimization utilizing the spatial context of BIM models. This process used Graph Neural Network-based (GNN) representations of the built environment context for a case study in Singapore to compare three scenarios of targeting samples from occupants being asked to give intensive longitudinal feedback. The results showed that the proposed novel representations of the spatial context mainly had higher degrees of sampling quality than the two baselines. The removal of redundant occupant feedback data points is similar in the performance of the baselines but with a higher level of scalability potential. The adaptive sampling quality can be further improved by implementing it online in real-time scenarios as a true digital twin application. This method can be used to significantly optimize subjective feedback data sampling by targeting occupants based on their location, personal, and environmental data. 

\subsection{Reproducibility}
Segments of the raw data and analysis code used for this study are available in an open-access Github repository that includes further documentation:
\href{https://github.com/buds-lab/build2vec1.0}{https://github.com/buds-lab/build2vec1.0}

\section*{CRediT author statement}
\textbf{MA:} Conceptualization, Methodology, Validation, Formal analysis, Writing - Original Draft, and Visualization, Data Curation, and Investigation. \textbf{CM:} Conceptualization, Investigation, Resources, Data Curation, Writing - Review \& Editing, Supervision, Project administration, and Funding acquisition.

\section*{Funding}
Singapore Ministry of Education (MOE) Tier 1 Grants provided support for the development and implementation of this research under the projects Ecological Momentary Assessment (EMA) for Built Environment Research (R296000210114) and The Internet-of-Buildings (IoB) Platform – Visual Analytics for AI Technologies towards a Well and Green Built Environment (R296000214114).

\section*{Acknowledgements}
The authors would like to acknowledge the team behind the dataset collection and processing, including Mario Frei, Matias Quintana, Yi Ting Teo, Yun Xuan Chua, Charlene Tan, Pearlyn Ang, Jin Kai Leong, Charis Boey, and Muhammad Zikry. The authors also would like to thank the SinBerBEST2 Lab, specifically Stefano Schiavon and Federico Tartarini, for contributing sensors and feedback. Also, we acknowledge the input and the best-practice guidelines provided by Prageeth Jayathissa during the experiment deployment. Finally, the authors would like to thank the National University of Singapore (NUS) for facilitating the deployment of sensors and BLE beacons across its buildings. This work contributes to the IEA EBC - Annex 79 - Occupant-Centric Building Design and Operation.

\bibliographystyle{model1-num-names}
\bibliography{references}

\end{document}